# Intelligent Robotic Control System Based on Computer Vision Technology

**Chang Che[1.1]***

1.1*Mechanical Engineering, The George Washington University, DC, USA

*Corresponding author:cche57@gwmail.gwu.edu,

**Haotian Zheng[1.2]**

1.2 Computer Engineering,New York University, New York, NY, USA

hz2687@nyu.edu

**Zengyi Huang[2.1]**

2.1 Applied Economics, The George Washington University, DC,USA

zengyihuang@gwmail.gwu.edu

**Wei Jiang [2.2]**

2.2 Computer Science, Xidian University, Xian,China

jiangwei13@gmail.com

**Bo Liu [3]**

3 Software Engineering, Zhejiang University ,Hangzhou China

lubyliu45@gmail.com

**Abstract.**

Computer vision is a kind of simulation of biological vision using computers and related equipment. It is an important part of the field of artificial intelligence. Its research goal is to make computers have the ability to recognize three-dimensional environmental information through two-dimensional images. Computer vision is based on image processing technology, signal processing technology, probability statistical analysis, computational geometry, neural network, machine learning theory and computer information processing technology, through computer analysis and processing of visual information.The article explores the intersection of computer vision technology and robotic control, highlighting its importance in various fields such as industrial automation, healthcare, and environmental protection. Computer vision technology, which simulates human visual observation, plays a crucial role in enabling robots to perceive and understand their surroundings, leading to advancements in tasks like autonomous navigation, object recognition, and waste management. By integrating computer vision with robot control, robots gain the ability to interact intelligently with their environment, improving efficiency, quality, and environmental sustainability. The article also discusses methodologies for developing intelligent garbage sorting robots, emphasizing the application of computer vision image recognition, feature extraction, and reinforcement learning techniques. Overall, the integration of computer vision technology with robot control holds promise for enhancing human-computer interaction, intelligent manufacturing, and environmental protection efforts.

# 1. Introduction

Computer vision technology is the process of simulating human visual observation and using computers to analyze images. It requires the computer to have the ability to perceive the surrounding environment through images and simulate the specific process of human vision to realize the intelligent processing of related images. Computer vision technology is a branch of artificial intelligence that mimics human perception of the environment. As a result, the technology integrates multiple disciplines and technologies, including image processing, artificial intelligence, and digital technologies.

In recent years, the integration of computer vision technology with robot control and interaction has opened up new avenues of research and development. Researchers are gradually realizing that vision is very important to people, [1]and more than 90% of human information is based on the eyes. Let's look at the frontier of artificial intelligence - machine vision. By equipping robots with computer vision capabilities, they gain the ability to perceive and interpret their surroundings, allowing them to interact intelligently with the environment and humans. The convergence of computer vision and robotics has led to significant advances in various fields, including autonomous navigation, object recognition, human-computer interaction, and industrial automation. With the development of the Internet of Things, there will be more and more precision industrial robots with multi-sensor and distributed control, which will gradually penetrate [2]all aspects of the manufacturing industry and transform from manufacturing implementation to service. Industrial robots with touch, force, or vision that can work in more complex environments; If they have the recognition function or further increase the adaptive and self-learning function, they become intelligent industrial robots.

Robotic vision, a rapidly growing branch of AI (artificial intelligence) that aims to give robots vision comparable to our own, has made huge strides in the past few years thanks to researchers applying specialized neural networks to help robots recognize and understand images from the real world. For example, in manufacturing environments, [3]robots equipped with computer vision can accurately identify and manipulate objects on assembly lines, increasing efficiency and productivity. In healthcare, robots equipped with computer vision can assist medical professionals with tasks such as surgery and patient care, increasing accuracy and reducing the risk of errors. In addition, the integration of computer vision technology with robot control enables robots to adapt to dynamic and unstructured environments, such as outdoor environments or disaster scenarios. By constantly analyzing visual data from the surrounding environment, the robot can make decisions in real-time and adjust its behavior accordingly, increasing its autonomy and versatility.

In this article, we will explore the intersection of computer vision technology and robotic control, examining the latest advances, applications, and challenges in this rapidly evolving field. We will delve into how computer vision algorithms are integrated into robotic systems, the impact and advantages of such integration for specific industries, and the prospects of computer vision technology in the field of machine vision interaction and control.

## 2. Related work

### 2.1. Computer Vision

Computer Vision, also known as Machine Vision, is a discipline that lets machines learn how to "see" and is an important application field of deep learning technology, which is widely used in security, industrial quality inspection and automatic driving scenarios. Specifically, it is to let the machine to identify the object in the picture or video taken by the camera, detect the location of the object, and track the target object, so as to understand and describe the scene and story in the picture or video, in order to simulate the human brain visual system. Therefore, computer vision is also commonly referred to as machine vision, and the goal is to build artificial systems that can "sense" information from images or videos.

The development of computer vision begins with biological vision. For the origin of biological vision, the academic community has not yet formed a conclusion. Some researchers believe that the earliest biological vision was formed in jellyfish about 700 million years ago, and others believe that biological vision emerged in the Cambrian period about 500 million years ago. [1][2] The cause of the Cambrian explosion has long been a mystery, but it is certain that with the advent of vision, predators were able to spot prey more easily and prey were able to locate predators earlier. The ability to see intensifies the game [4]between hunter and prey and gives rise to more intense evolutionary rules for survival. The formation of the visual system has strongly promoted the evolution of the food chain and accelerated the process of biological evolution, which is an important milestone in the history of biological development. After hundreds of millions of years of evolution, the human visual system has a very high complexity and powerful function, the number of neurons in the human brain has reached 100 billion, these neurons are connected through the network, such a huge visual neural network allows us to easily observe the world around.

Computer vision tasks depend on image features (image information), and the quality of image features largely determines the performance of the vision system. Traditional methods usually use SIFT, HOG and other algorithms to extract image features, and then use SVM and other machine learning algorithms to further process these features to solve visual tasks. Pedestrian detection is to determine whether there are pedestrians in the image or video sequence and give accurate positioning. The earliest method is HOG feature extraction +SVM classifier, and the detection process is as follows:

- The sliding window is used to traverse the whole image, and the candidate region is obtained
- Extract the HOG feature of the candidate region
- Classification of feature maps using SVM classifiers (to determine whether it is human)
- Using the sliding window will appear duplicate areas, using NMS(non-maximum) to filter the duplicate areas

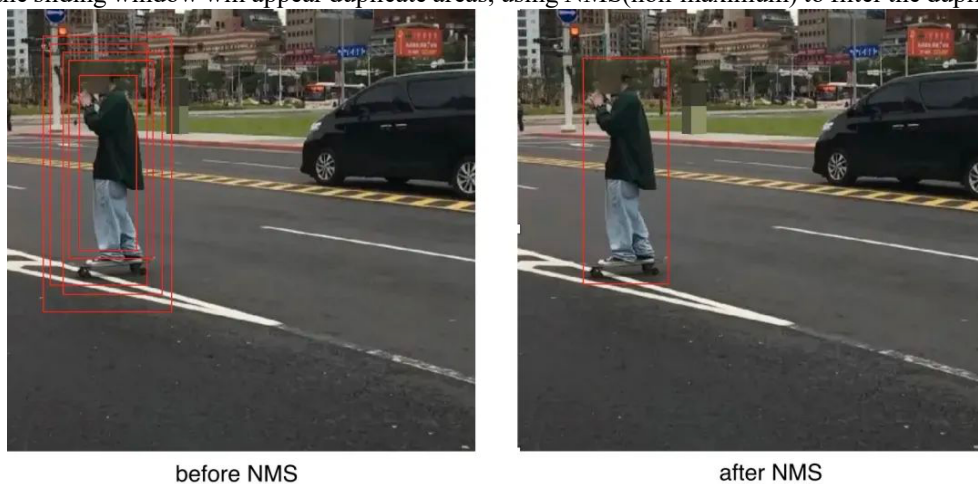

**Figure 1.** Examples of computer image detection

The application of machine vision in robot interaction and control is very broad, it can help robots perceive and understand the surrounding environment, so as to perform tasks and interact with humans more effectively. Below I will explain how the main ideas of machine vision can be combined with robot interaction and control.

Image processing stage:

In the robot vision system, the image processing stage is very important. [5]The robot acquires image data from the environment through cameras or other sensors.

In the process of image processing, the robot uses algorithms and technologies to preprocess the image, such as removing noise, enhancing contrast, edge detection, etc., in order to better extract the features of the object.

1. Object feature screening:

In the process of image processing, the robot needs to recognize and extract various physical characteristics of the object, such as shape, color, texture, etc.

The screening process also involves removing disturbing factors from the image, making the final feature more realistic and effective.

2. Image recognition stage:

After image processing, the robot performs the image recognition phase. This phase involves matching and analyzing the features of the screened object with data from a pre-established database.

By identifying and matching the features of the object, the robot can determine the identity, location and state of the object.

3. Robot interaction and control:

Once a robot recognizes objects in its surroundings, it can perform various tasks and interactive behaviors based on this information.

For example, robots can navigate, grasp, control devices, and communicate with humans based on the objects they recognize.

4. Real-time feedback and adjustment:

During the execution of tasks, the robot may need to constantly make real-time feedback and adjustments. This can be achieved by continuously acquiring image data and processing it.

In summary, [6]with the continuous development of intelligent robots in the future, people's requirements for human-computer interaction are becoming higher and higher, intelligent, smooth, and anthropomorphic, which deeply test our ability to apply various machine modules. Here we discuss the human-machine interaction between machine vision and artificial intelligence from three aspects: robot vision artificial intelligence and robot control.

## 2.2. Robot visual interaction and control

Machine vision encompasses various technologies, including image processing, lighting control, optical sensors, and computer software, aimed at enhancing production flexibility and automation. It finds applications where manual labor may be impractical, substituting it with automated processes driven by machine vision. Integrating machine vision with robotics and artificial intelligence (AI) is crucial for advancing human-computer interaction capabilities.

In industrial automation, machine vision systems empower robots to perform precise tasks on production lines. [7]By accurately identifying and positioning objects, robots execute assembly, inspection, and packaging tasks, thus boosting production efficiency and quality. Similarly, in logistics and warehousing, robot vision enables autonomous navigation, cargo identification, and handling, leading to enhanced operational efficiency, reduced labor costs, and fewer errors.The medical field also benefits from robot vision technology, particularly in surgical procedures. Advanced computer vision systems assist surgeons in performing precise operations, minimizing trauma and recovery time. Moreover, robot vision aids in medical image analysis, disease diagnosis, and rehabilitation treatment, offering innovative solutions for healthcare.

The ultimate goal of machine vision is to mimic human recognition abilities. Technological advancements, including recognition algorithms and hardware support like light sensors and image processing units, enhance machine vision's capability to recognize objects and faces intelligently. By optimizing algorithms and increasing recognition efficiency, machine vision systems become more adept at identifying objects in real-world scenarios.Real-world applications of machine vision technology bring convenience and efficiency to people's lives. For instance, in manufacturing, automation driven by machine vision reduces production time and enhances product quality. In logistics, automated systems powered by machine vision streamline operations, ensuring timely and accurate delivery of goods. Moreover, in healthcare, robot vision aids medical professionals in delivering precise treatments and diagnoses, ultimately improving patient outcomes.

*2.3. Robot visual interaction application*

Human-computer interaction (HCI) [8]has become ubiquitous in various aspects of daily life, from simple radio buttons to complex control systems in critical environments like nuclear reactors. With the rapid advancement of technology, machine vision has emerged as a crucial element in HCI, offering new possibilities and dimensions to the interaction process.Traditionally, machine vision and robot control have operated in decentralized manners, each focusing on specific tasks or domains. For example, machine vision is often used for sample screening in manufacturing to identify defective products, while remote-controlled robots handle high-risk tasks under human supervision. However, these approaches often lack sophistication, leading to limitations in HCI.

The advent of intelligent robots has transformed this landscape by integrating built-in vision sensors, robotic frameworks, and rudimentary logical reasoning capabilities. These robots can now engage in meaningful communication and interaction with humans. For instance, systems like AlphaGo leverage [9]machine vision, artificial intelligence, and deep learning to facilitate HCI with human players, showcasing the potential of advanced technologies in enabling seamless human-robot interaction.Machine vision plays a crucial role in HCI applications by enabling intelligent robots to perceive and understand their surroundings, recognize objects, and perform tasks autonomously. For example, smart home robots equipped with machine vision technology can identify objects within the household and execute cleaning tasks. They can also utilize facial recognition to differentiate family members and provide personalized services.

Despite the fragmented integration between machine vision and robot control in traditional production processes, ongoing advancements in science and technology are driving improvements in HCI capabilities. The emergence of intelligent robots equipped with visual sensors and mechanical frameworks signifies a shift towards robots possessing rudimentary logical thinking abilities and the capacity for human-like interactions.

As an illustrative example, consider the task of waste sorting, a common chore in households and communities. Traditional waste sorting methods often rely on manual inspection and sorting by humans, which can be time-consuming and prone to errors. In contrast, intelligent robots equipped with machine vision technology can accurately identify different types of waste materials through image recognition, streamlining the sorting process and improving efficiency. Additionally, these robots can adapt to changing environments and learn from experience, further enhancing their effectiveness in waste management tasks.

The integration of machine vision and artificial intelligence in intelligent robots holds significant promise for enhancing HCI applications. By leveraging advanced technologies, these robots offer greater convenience and efficiency in various aspects of human life, while driving ongoing progress and innovation in HCI technology.

## 3. Methodology

With the acceleration of global urbanization, urban waste disposal has become an increasingly serious problem.  Traditional garbage sorting methods usually rely on manual sorting, which is inefficient and error-prone.  To further explore the intersection of computer vision technology and robot control, it is imperative to establish a comprehensive methodology that encompasses various aspects of research, development, and application.  Therefore, this part using robot vision technology to develop intelligent garbage sorting robots has become a new way to solve this problem

*3.1. Application design*

Above the garbage collection and compression bin of the garbage transfer station, an intelligent identification camera is equipped, which can recognize the license plate of the garbage truck with high precision, and carry out real-time visual detection and recognition of the dynamically dumped garbage. By using advanced algorithm deduction technology, the recognition results are compared with the graph database of garbage, so as to realize the high-precision detection and recognition of garbage. And through the early warning system to remind the relevant personnel to deal with.

This device has the functions of real-time monitoring, automatic identification, quick warning and so on, which can effectively improve the efficiency of waste treatment, save manpower and reduce equipment loss. Therefore, through the visual interaction and control of intelligent garbage sorting robot, the robot vision technology can be used to realize the automatic identification and classification of garbage. The robot can accurately identify various types of garbage in the garbage disposal station or household and other scenes, and place them respectively in the corresponding recycling bins, thereby improving the efficiency of garbage disposal and reducing environmental pollution.

*3.2. Implementation principle*

1) Computer vision image recognition, image recognition is an important research direction in the field of artificial intelligence and computer vision. Its development is inseparable from the progress of computer technology and the support of big data. [10]At present, image recognition has been widely used in face recognition, unmanned driving, intelligent security monitoring, medical image analysis, virtual reality and other fields. Computer vision algorithm is the core of image recognition, which uses mathematical and statistical methods to extract features from images, and finally realizes image classification, detection and recognition after model training and optimization.

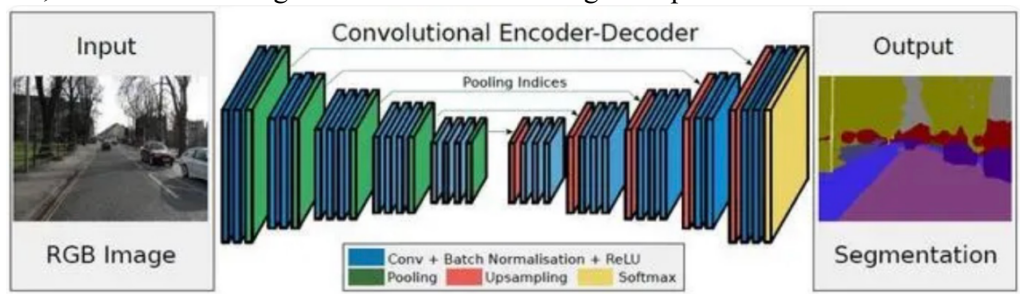

**Figure 1.** Principle of computer vision image recognition

2) Image recognition method

**Feature extraction:** Feature extraction is the primary task of computer vision algorithms. It extracts representative feature information, such as color, texture, shape, etc., by processing pixels in the image. Commonly used feature extraction methods include SIFT, SURF, HOG, etc.

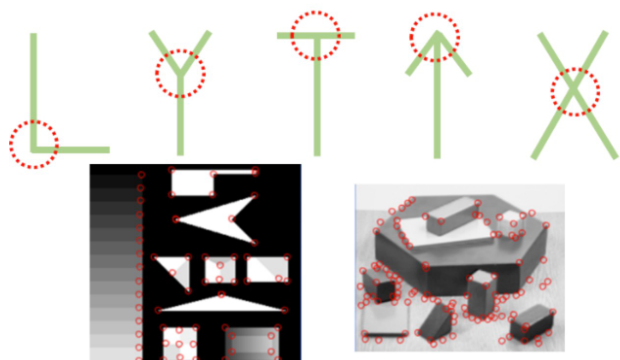

**Figure 2.** Principle of image feature extraction

**Model training and optimization:** On the basis of feature extraction, it is necessary to build a suitable model to realize image classification and recognition. Commonly used models include support vector machine (SVM), K-nearest neighbor (KNN), deep neural network (DNN), etc. The process of training the model is to learn and optimize through a large number of image samples, so that the model can accurately classify and recognize images.

**Target detection and recognition:** Target detection and recognition is one of the core tasks of image recognition. It is able to accurately find out the target object from the image and identify it according to the model established in advance. Target detection and recognition methods mainly include YOLO (You Only Look Once), Faster R-CNN, and SSD (Single Shot MultiBox Detector) based on deep learning.

Therefore, this application experiment is closely related to image recognition and feature extraction through the application of machine vision in intelligent garbage classification. First, image recognition technology enables the computer to identify the type of waste, such as paper, plastic, metal, etc., which helps the classification system understand the input image. [11]Second, feature extraction refers to the process of extracting useful information from images, such as color, texture, shape, etc. These features help the sorting system distinguish between different types of garbage. Therefore, the machine vision system needs to combine image recognition technology to identify the type of garbage, and use feature extraction technology to extract the features in the image to achieve accurate garbage classification.

### 3.3. Robot operation and motion control

In a smart waste sorting experiment in which computer vision technology interacts with an intelligent robot, the researchers conducted experiments using a reinforcement learning system in robot classrooms. The researchers collected a large amount of trial data, both in the actual deployment environment and the simulation environment. By continuously adding data, they improved the performance of the entire system and evaluated the final system. The experimental results show that the system has achieved good results in handling garbage sorting tasks, with an average accuracy of about 84%, and the performance has steadily improved with the increase of data.

In practical applications, the researchers also recorded statistics on the actual deployment of the system from 2021 to 2022, and found that the system was effective in reducing pollutants in trash cans. By combining offline and online data, the robot is able to adapt to a variety of situation changes in the real world, demonstrating the potential of systems based on reinforcement learning when dealing with real tasks.

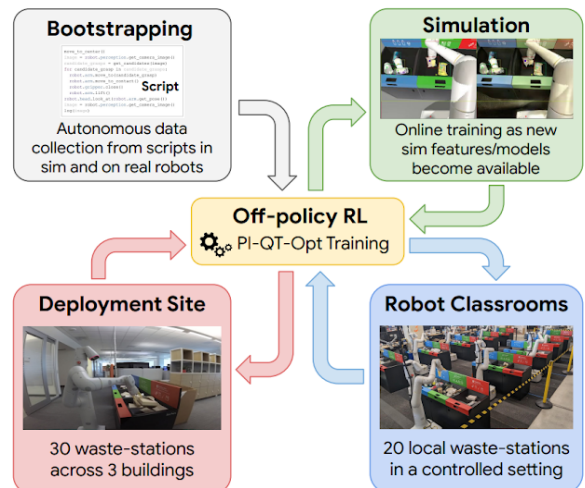

**Figure 3.** Schematic diagram of reinforcement learning in this large-scale application to achieve garbage classification and detection

The system presented in Figure 3 is based on the QT-Opt reinforcement learning framework, which has the advantage of being able to effectively guide the robot to learn a variety of tasks, including grasping of different garbage and other skills. In a laboratory setting, by using the framework, the robot can start learning from simple scripting strategies, gradually transition to reinforcement learning, and combine the Cyclegan-based migration method and RetinaGAN technology to make the simulation images more realistic.

Computer vision plays an important role in this system, providing critical information and feedback to the robot. Through computer vision, the robot is able to recognize the type and location of different garbage, so that it can effectively grasp and dispose of it. In addition, with RetinaGAN technology, the fidelity of the simulation images is improved, enabling the robot to better learn in the simulation environment and apply what it learns in the actual deployment environment.

This method of integrating computer vision technology provides strong support for robots to perform tasks in complex environments, and also provides a good foundation for them to adapt and learn in real scenarios.

### 3.4. Robot motion control and real-time feedback

In practice, an intelligent robot that actually achieves a high degree of motion control and real-time feedback through computer machines is [12]AMP Robotics, headquartered in Denver, Colorado. On the morning of January 4, AMP Robotics announced that it had closed a $55 million funding round. AMP Robotics said it will use the funding to scale operations, develop applications for artificial intelligence products, and integrate them into waste stream recycling to improve recycling rates for customers. At present, automation is expected to drive up the total amount of waste recycling at a time when manual recycling is facing inefficiencies, and because of this, the market for garbage sorting robots is expected to reach $12.26 billion by 2024 and continue to grow at a compound annual growth rate of 16.52%.

The AMP Robotics garbage sorting robot has an unusually high sorting speed - 80 pieces per minute, much higher than manual pickup, and can automatically identify, process and monitor complex waste streams. It is modular in design, allowing managers to adapt existing workflows for a single class or different volumes of recyclable objects. AMP robots can pick out many kinds of garbage, including metal, batteries, capacitors, plastics, [13]PCBS, wires, cartons, cardboard, cups, shells, LIDS, aluminum, etc. It also includes metal blends of wood, asphalt, brick, concrete, plastics and mixtures (e.g., polyethylene terephthalate, high density polyethylene, low density polyethylene, polypropylene, polystyrene). It also classifies the film by color, clarity and transparency.

AMP Robotics realizes the workflow of action control and real-time feedback for garbage classification as follows:

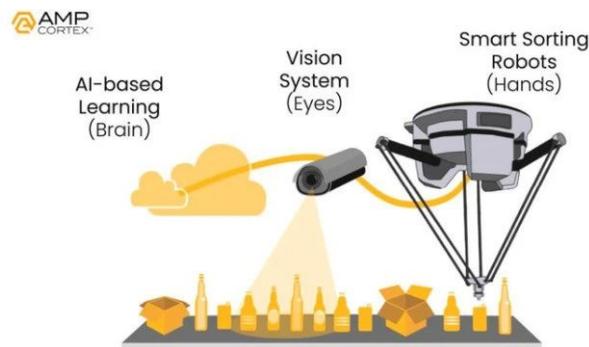

**Figure 4.** Robot garbage sorting principle architecture released by AMP company

As can be seen from Figure 4, in the process of garbage classification, the robot carries out the classification business through the following processes combined with computer vision technology:

1.Garbage delivery and delivery: Garbage is dropped by the customer into the garbage collection scene, and then transported to the robot's work area by conveyor belt or other means.

2.Visual recognition and localization: AMP Robotics' robots are equipped with an advanced visual recognition system to quickly and accurately identify various objects in the garbage stream. Once the target object is identified, the robot immediately locates its position.

3.Motion control and grasping: According to the visual recognition and positioning results, the robot quickly adjusts its posture and executes corresponding actions through its own motion control system to accurately grasp the target object.

Real-time feedback and adjustment: AMP Robotics' robots are equipped with a real-time feedback system to continuously monitor various parameters in the garbage sorting process, such as the capture success rate and classification accuracy rate. Once there is an abnormal situation or wrong classification, the system will immediately give feedback and automatically adjust according to the situation.

The training data of AMP Robotics comes from AMP Neuron, and the built-in algorithm of the robot trains through massive images, including various recyclable wastes such as water bottles, beer cans, milk cans, and food packaging boxes. The image database contains the various states in which the recyclables are intact, indentation or crushed. This allows robots to classify more accurately, learn new material classes more efficiently, and better adapt to packaging design and lighting changes. Sensors use computer vision to scan fast-moving objects, distinguishing them based on color, texture, shape, size, and material. [14]The robot's attachments then use suction cups to grab items from the conveyor belt and drop them into the corresponding recycling bin.

*3.5.    Intelligent robot visual inspection interaction advantages*

1. Real-time monitoring and automatic recognition: The garbage visual identification and warning device uses high-definition cameras to monitor the garbage stacking area in real time and collect a large number of image data. Through deep learning algorithms, such as convolutional neural networks (CNN), image preprocessing, feature extraction and classification are carried out to judge the garbage style and category. This real-time monitoring and automatic identification technology can greatly improve the efficiency and accuracy of waste disposal.

Rapid warning and notification: Once the device recognizes the presence of large garbage in the garbage dump, it will immediately issue an early warning signal to notify the relevant personnel to deal with it. This fast warning system can prevent the damage of large garbage to garbage disposal equipment in time, and improve the overall efficiency and safety of garbage disposal.

Saving manpower and reducing costs: The large waste visual identification early warning device realizes automatic monitoring and processing, reducing the workload of manual inspection, thus reducing labor costs. At the same time, by collecting a large amount of waste disposal data, and conducting data analysis and mining, it can provide optimization suggestions for waste disposal work, and further reduce the overall operating cost.

2. Data analysis and feedback: By comparing the data to analyze the classification of all incoming waste, including collecting the total volume data of garbage, the total volume data of illegal garbage and the license plate data of illegal garbage, etc., the garbage transfer and compression work can be better managed to avoid more illegal large garbage entering the garbage disposal transfer station. This data-driven feedback mechanism helps to continuously improve the efficiency and quality of waste disposal efforts.

3. Environmental protection and sustainable development: The operation of the visual identification and early warning device for large garbage can deal with large garbage in time, thereby reducing the waste of land resources and environmental pollution caused by garbage accumulation. [15-17]At the same time, because the device does not need to consume a lot of energy during operation, it has a low carbon emission, which is conducive to environmental protection and sustainable development. In addition, the device can also be combined with other environmental protection equipment, such as garbage sorting equipment, recycling equipment, etc., to jointly build a more green and environmentally friendly ecosystem.

In summary, the methodology for developing intelligent garbage sorting robots through the integration of computer vision technology and robot control offers a comprehensive approach to address urban waste management challenges. By equipping garbage transfer stations with intelligent identification cameras and advanced algorithm deduction technology, these systems can achieve high-precision garbage detection and recognition, improving waste treatment efficiency and reducing manual labor.

Implementation principles such as computer vision image recognition, feature extraction, and model training enable accurate garbage classification and detection. Reinforcement learning techniques further enhance the capabilities of intelligent robots in handling garbage sorting tasks, both in simulated and real-world environments. Through these advancements, the integration of computer vision technology into robotic systems offers practical solutions for improving waste management processes and contributing to environmental sustainability.

## 4. Conclusion

A robot is a machine that can be programmed and automatically controlled to perform tasks such as jobs or movements. In recent years, the global robot market has continued to grow, from $26.7 billion in 2017 to $51.3 billion in 2022[18], and is expected to continue to grow to $66 billion by 2024. The visual identification and early warning robot device is widely used in traffic, public security, sanitation, administrative management, and other scenes of smart cities, which is of great significance in improving the efficiency of visual identification and resolution in public areas and saving public resources. Additionally, the visual recognition robot device also has strong scalability. According to actual demand, the recognition function of other types of waste can be added to realize the comprehensive management of domestic waste. It can also be linked with other smart devices, such as the intelligent switch of the garbage bin and the automatic sorting of the garbage sorting equipment, to further improve the efficiency of garbage disposal.

In general, this paper discusses the intersection of computer vision technology and robot control in depth, emphasising the importance and potential of this field in human-computer interaction, intelligent manufacturing and environmental protection. This paper first introduces the definition and development of computer vision technology, points out its important role in simulating human visual observation and image analysis, and emphasises its importance in human information acquisition. Then, the paper introduces the integration of computer vision technology and robot control in detail, as well as the influence and application of this integration in various fields. Especially in the fields of industrial automation, healthcare and environmental protection, this integration has brought great progress and benefits, improving production efficiency, quality and environmental protection.